\documentclass[conference]{IEEEtran}
\IEEEoverridecommandlockouts
\usepackage{cite}
\usepackage{amsmath,amssymb,amsfonts}
\usepackage{algorithmicx}
\usepackage{graphicx}
\usepackage{textcomp}
\usepackage{multirow} 
\usepackage{booktabs}
\usepackage{url}
\usepackage{booktabs}       

\usepackage{amsfonts,amsmath}       
\usepackage{nicefrac}       
\usepackage{microtype}      

\usepackage{multirow}       
\usepackage{graphics}
\usepackage{tabularx}
\usepackage{mathrsfs}
\usepackage{makecell}
\usepackage{caption}
\usepackage{wrapfig}
\usepackage{enumitem}
\usepackage{subcaption}
\usepackage{bbding}

\usepackage{algpseudocode}
\usepackage{amsfonts}       
\usepackage{color, colortbl}
\usepackage{xcolor}
\definecolor{Gray}{gray}{0.95}
\definecolor{orange}{rgb}{0.9,0.5,0}
\definecolor{LightCyan}{rgb}{0.88,1,1}
\def\BibTeX{{\rm B\kern-.05em{\sc i\kern-.025em b}\kern-.08em
    T\kern-.1667em\lower.7ex\hbox{E}\kern-.125emX}} 
\definecolor{DeepBlack}{RGB}{10, 10, 10}
\usepackage[colorlinks,citecolor=blue,linkcolor=red,bookmarks=false]{hyperref}

\begin{document}

\title{Learning Dual-Domain Multi-Scale Representations for Single Image Deraining}

\author{
	\IEEEauthorblockN{
		Shun Zou\IEEEauthorrefmark{1}$^{\ast}$\thanks{$^{\ast}$Equal Contribution, $^{\dagger}$Corresponding author.}, 
		Yi Zou\IEEEauthorrefmark{2}$^{\ast}$, 
        Mingya Zhang\IEEEauthorrefmark{3},
		Shipeng Luo\IEEEauthorrefmark{4},
        Guangwei Gao\IEEEauthorrefmark{5}$^{\dagger}$,
        Guojun Qi\IEEEauthorrefmark{6}} 
        
        \IEEEauthorblockA{%
      \begin{tabular}{cccc}
        \IEEEauthorrefmark{1}Nanjing Agricultural University & \IEEEauthorrefmark{2}Xiangtan University &
        \IEEEauthorrefmark{3}Nanjing University &
        \IEEEauthorrefmark{4}Northeast Forestry University 
      \end{tabular}%
    }
    
    \IEEEauthorblockA{%
  \begin{tabular}{cc}
    \IEEEauthorrefmark{5}Nanjing University of Posts and Telecommunications &
        \IEEEauthorrefmark{6}Westlake University
  \end{tabular}%
}

	\IEEEauthorblockA{
\href{https://zs1314.github.io/DMSR}{https://zs1314.github.io/DMSR}}
}

\DeclareRobustCommand*{\IEEEauthorrefmark}[1]{%
    \raisebox{0pt}[0pt][0pt]{\textsuperscript{\footnotesize\ensuremath{#1}}}}

\maketitle

\begin{abstract}
Existing image deraining methods typically rely on single-input, single-output, and single-scale architectures, which overlook the joint multi-scale information between external and internal features. Furthermore, single-domain representations are often too restrictive, limiting their ability to handle the complexities of real-world rain scenarios. To address these challenges, we propose a novel Dual-Domain Multi-Scale Representation Network (DMSR). The key idea is to exploit joint multi-scale representations from both external and internal domains in parallel while leveraging the strengths of both spatial and frequency domains to capture more comprehensive properties. Specifically, our method consists of two main components: the Multi-Scale Progressive Spatial Refinement Module (MPSRM) and the Frequency Domain Scale Mixer (FDSM). The MPSRM enables the interaction and coupling of multi-scale expert information within the internal domain using a hierarchical modulation and fusion strategy. The FDSM extracts multi-scale local information in the spatial domain, while also modeling global dependencies in the frequency domain. Extensive experiments show that our model achieves state-of-the-art performance across six benchmark datasets.
\end{abstract}


\section{Introduction}
\label{sec:intro}
Single-image deraining (SID) is a vital task in low-level vision, focused on restoring clear background images from rainy inputs by eliminating or reducing undesirable degradations caused by rain artifacts \cite{chen2024rethinking}. The severe interference of rainy images with the performance of downstream tasks has sparked significant research interest in the SID task.

Over the years, numerous methods have been proposed to address the image deraining problem. Early prior-based traditional approaches \cite{Xu2012AnIG,zhang2017convolutional} attempted to remove rain artifacts by analyzing the statistical characteristics of rain streaks and the background. However, in real-world scenarios, rain streaks and droplets are often dense, complex, and diverse, causing these methods to fail in such cases. Recently, many CNN-based methods \cite{li2018recurrent,ren2019progressive,Chen_2021_CVPR,cui2023exploring,cui2024omni} have been applied to image deraining, bringing transformative advancements to the field. Compared to traditional algorithms, these methods have significantly improved deraining results. However, convolution, as the core element of CNNs, exhibits spatial invariance and a limited receptive field, making it inadequate for effectively modeling the non-local structures and spatial variations in clear images \cite{song2024learning}. To address these limitations, some methods \cite{Chen_2023_CVPR,chen2023hybrid,SFHformer,zhou2024AST,NeRD-Rain} have adopted Transformers for image deraining, achieving impressive results. Transformers effectively capture global dependencies and model non-local information, enabling high-quality image reconstruction.

However, existing mainstream supervised paradigms still face two major challenges: (1) the inability to explore complementary implicit information across different scales. Most current deep deraining networks rely on single-input single-output architectures. However, spatially varying rain streaks often exhibit diverse scale attributes, which not only overlooks potential explicit information from multiple image scales but also fails to explore complementary implicit information across scales. Generally, multi-scale visual information flow involves representations from external interactions (multi-input multi-output exchanges with the network’s external environment) and internal interactions (scale information exchange within internal components of the network). Some researchers have introduced coarse-to-fine mechanisms \cite{xintm2023DeepRFTv2,chen2024rethinking,Kim2022MSSNet} or multi-patch strategies \cite{Zamir2021MPRNet,9157139} to leverage multi-scale external rain features. While these methods have achieved impressive performance, they struggle to handle complex and random rain streaks because they rarely consider collaborative multi-scale feature representations from both external and internal domains, neglecting implicit relationships across scales. (2) Rain streaks and droplets consistently exhibit irregular overlaps and highly dynamic geometric patterns, which place higher and more diverse demands on feature representations. Moreover, in real-world rainy images, significant aliasing effects exist between rain residues and the background. Eliminating rain disturbances by inferring pixel residue values inevitably compromises contextual and structural information. When features are derived solely from a single domain or dimension, it becomes challenging to remove all interferences \cite{cui2024dual}. Therefore, more robust and diverse multi-domain feature representations are critical for achieving high-quality image reconstruction.

To address the abovementioned challenges, we introduce DMSR, a novel dual-domain multi-scale architecture model. It consists of multiple Dual-Domain Scale-Aware Modules (DDSAM), which include two key components: the Multi-Scale Progressive Spatial Refinement Module (MPSRM) and the Frequency Domain Scale Mixer (FDSM). To tackle the first challenge, we propose a collaborative multi-scale paradigm for both external and internal domains. We use a coarse-to-fine pyramid input-output flow externally across the entire network while applying various multi-scale architectures within the internal components of DMSR (i.e., MPSRM and FDSM). This facilitates the flow of information within and across scales, enabling the interaction and fusion of complementary implicit information between different scales. Additionally, to address the second challenge, we use MPSRM to extract spatial domain features and introduce Spatial Pixel Guided Attention (SPGA) to aggregate information for each pixel, allowing each pixel to perceive information from an extended region implicitly. Simultaneously, FDSM decouples the features into distinct frequency domain components and modulates them, promoting interactions between different frequencies and refining the spectrum. This adaptive extraction and refinement of rain residuals and background components allows us to fully capture information from different dimensions through the extraction of dual-domain features, facilitating the exchange and complement of expert knowledge.
Our contributions are summarized as follows:  
\begin{itemize}
\item We propose a novel dual-domain multi-scale architecture that rethinks the multi-scale representation paradigm for single-image deraining through a collaborative multi-scale paradigm across internal and external domains, enabling better extraction of rich scale-space features. At the same time, it fully leverages the advantages of the dual-domain paradigm to achieve more diverse, multi-dimensional, and robust high-level feature representations. 
\item We introduce a Multi-Scale Progressive Spatial Refinement Module for extracting refined spatial pixel features, using gated contextual information to integrate implicit information from the surrounding context and expand the receptive field. Additionally, we develop a Multi-Scale Frequency Domain Mixer that combines global and local characteristics and facilitates information exchange and modulation across different spectral components, thereby generating high-quality deraining results. 
\end{itemize}

\section{Method}
\subsection{Overall Pipeline}
As shown in Fig. \ref{overview} \textcolor{red}{(a)}, DMSR is divided into three scales based on a coarse-to-fine approach. Specifically, given an input rain image, the original image is downsampled by an interpolation operator to 1/2 and 1/4 of the original size, forming a pyramid of rain image inputs, with the coarsest to finest scale inputs referred to as S1, S2, and S3. We first pass S1 through a \( 3 \times 3 \) convolutional layer to obtain shallow features \( F \in \mathbb{R}^{C \times H \times W} \), where C, H, and W represent the channels, height, and width, respectively. These shallow features are then processed through three DDSAM to obtain multi-scale high-level representation features in both spatial and frequency domains. As shown in Fig. \ref{overview} \textcolor{red}{(b)}, DDSAM consists of multiple residual structures and includes the MPSRM and the FDSM. During this process, the spatial resolution is reduced by half, and the channel number is doubled. 
Moreover, we incorporate the downsampled rain images S2 and S3 into the main path after passing them through an Embedding Layer and adjusting the channel number with a \( 3 \times 3 \) convolution. The dual-domain multi-scale high-level representation features are then passed through another three DDSAMs for gradual restoration into a high-resolution derained image. During this process, features from the encoder and decoder are concatenated to facilitate image restoration. In line with the multi-scale inputs, we also employ multi-scale outputs, generating low-resolution derained images after the first two DDSAMs in the restoration process. Finally, skip connections are applied across the three pyramid input/output image pairs. For simplicity, only the top-level skip connection between the rain image and the derained image is shown in Fig. \ref{overview} \textcolor{red}{(a)}.

\subsection{Multi-Scale Progressive Spatial Refinement Module}
\begin{figure*}[ht]
	\centering
	\includegraphics[width=\linewidth]{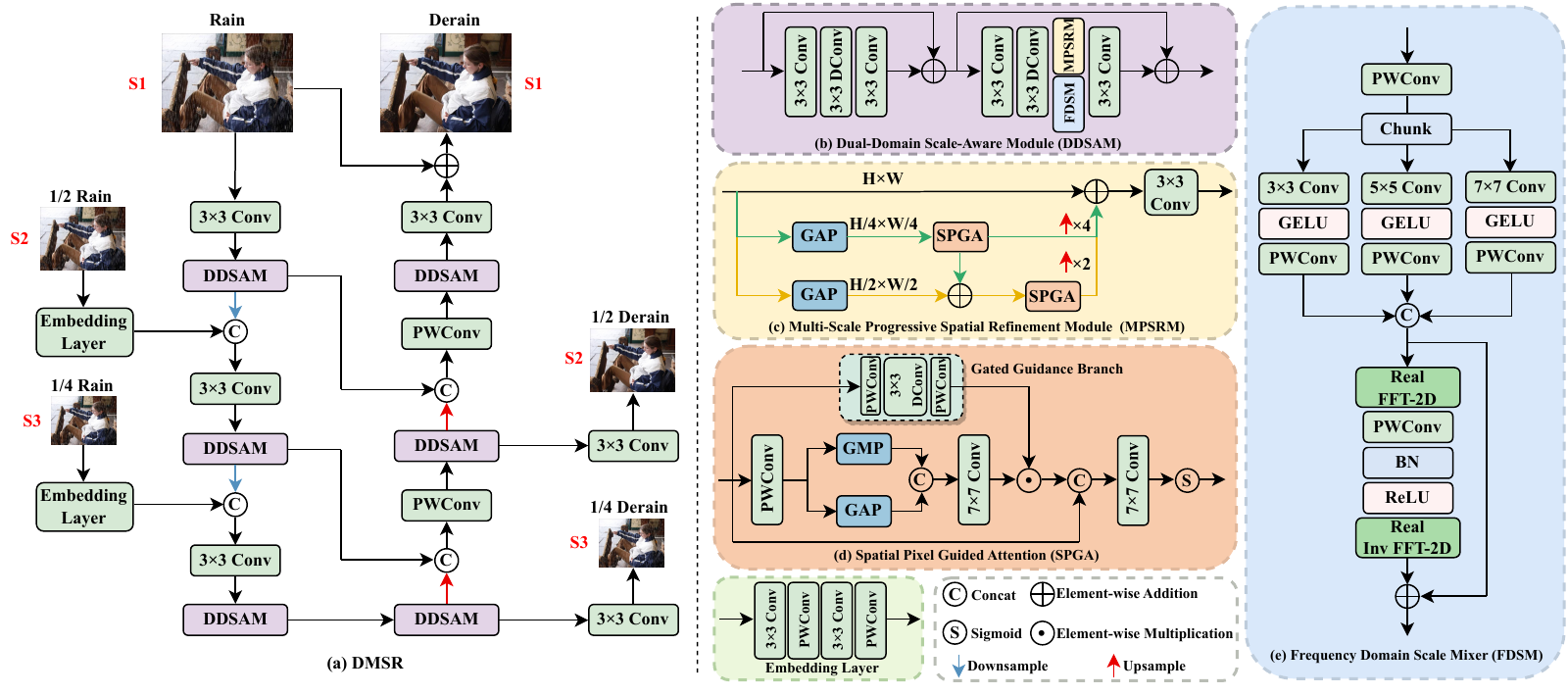} 
	\vspace{-7mm}
	\caption{The overall architecture of the proposed DMSR.}
	\label{overview}
	\vspace{-7mm}
\end{figure*}

The architecture of MPSRM is shown in Fig. \ref{overview} \textcolor{red}{(c)}. To alleviate the knowledge discrepancy across different scales within the same representation range, we perform high-quality restoration in a progressive coarse-to-fine manner within each representation scale, efficiently achieving hierarchical representation. Additionally, MPSRM focuses on modeling spatial context to enhance the spatial pixel representation ability of each feature map. By capturing pixel context relationships, it aids in the precise restoration of background and fine details. Specifically, given the input feature \(F\), global average pooling with different downsampling rates is applied to embed the initial feature \(F\) into different scales. For each scale (i.e., each branch), we input the scale features into the Spatial Pixel Guided Attention (SPGA), enhancing the expressiveness of each feature map. The resulting spatially refined features are then integrated into the next branch via addition. This enables MPSRM to progressively reduce intra-scale differences and promote the flow of expert information across different scales. Finally, the progressively fused features from all branches are unified to match the input size of \(F\) and summed. Formally, the above process is described as follows:
\vspace{-2mm}
\begin{equation}
    \hat{F_{1}} =SPGA(GAP_{4}(F)),
    \vspace{-2mm}
\end{equation}
\begin{equation}
    \hat{F_{2}} =SPGA(GAP_{2}(F)+\hat{F_{1}} ),
    \vspace{-2mm}
\end{equation}
\begin{equation}
    \hat{F} =f^{3\times 3}(F+\hat{F_{1}}\uparrow_4 + \hat{F_{2}}\uparrow_2 ),
    \vspace{-2mm}
\end{equation}
where \( GAP_i(\cdot) \) denotes global average pooling with a downsampling rate of \( i \), \( \uparrow_i \) represents bilinear upsampling with a scaling factor of \( i \), and \( f^{z \times z}(\cdot) \) indicates a convolution with a kernel size of \( z \times z \).

\subsection{Spatial Pixel Guided Attention}
In traditional spatial self-attention mechanisms \cite{dosovitskiy2020image}, spatial context modeling is often achieved by computing spatial feature distribution maps to indicate the importance levels of different regions. However, due to the complexity of real-world rainfall scenarios and the high interweaving of rain streaks with rain-free backgrounds, fine texture details are often disrupted during restoration, resulting in artifacts and noise. To address these issues, we propose a Spatial Pixel Guided Attention (SPGA) mechanism, which shifts from region-based to pixel-level guidance. By gradually generating perception information for each pixel through gated guidance, SPGA facilitates pixel-wise information interaction, emphasizing the most valuable expert information in feature maps and safeguarding texture details from contamination.
The structure of SPGA is illustrated in Fig. \ref{overview} \textcolor{red}{(d)}. Specifically, given the input feature \( X \in \mathbb{R}^{C \times H \times W} \), spatial attention is first applied to generate the spatial feature distribution map \( W_S \). Furthermore, we design a gated guidance module to focus features on the most valuable information. The gating mechanism promotes the flow of critical information while preventing excessive redundancy within intermediate layers, thereby mitigating information loss. This process can be mathematically described as: 
\vspace{-2mm}
\begin{equation}
    W_{S}=f^{7 \times 7}([GAP_{2}(PW(X)),GMP_{2}(PW(X))]),
    \vspace{-2mm}
\end{equation}
\begin{equation}
    W_{G}=PW(DW^{3\times 3}(PW(X))),
    \vspace{-2mm}
\end{equation}
where \( {PW}(\cdot) \) denotes point-wise convolution, \( {DW}^{3 \times 3}(\cdot) \) represents depth-wise convolution with a kernel size of \( 3 \times 3 \), \( [\ ,\ ] \) denotes concatenation, and \( {GMP}_i(\cdot) \) represents global max pooling with a downsampling rate of \( i \). Subsequently, we fuse \( W_S \) and \( W_G \) through an additive operation to obtain the coarse spatial pixel feature map \( W_M \). Then, we use a \( 7 \times 7 \) convolution operation to further expand the receptive field, re-weighting the pixel information to enable each pixel to perceive signals from a large region centered around itself, while employing the sigmoid function to introduce non-linearity. Formally, the entire process is expressed as follows:
\vspace{-2mm}
\begin{equation}
    W_{M}=W_{S}\odot W_{G},
    \vspace{-2mm}
\end{equation}
\begin{equation}
    W=\delta (f^{7\times 7}([X,W_{M}])),
    \vspace{-2mm}
\end{equation}
where \(\delta(\cdot)\) denotes the sigmoid activation function, \(\odot\) represents element-wise multiplication.  

\subsection{Frequency Domain Scale Mixer}
Fourier transform exhibits irreplaceable advantages in single image deraining. First, it can effectively separate image degradation components, as rain streak patterns display prominent and invariant characteristics in the frequency domain, which is crucial for mitigating aliasing effects in rainy images \cite{FADformer,SFHformer}. Second, the transformed frequency components, calculated from all spatial components, serve as global feature filters \cite{FSNet,cui2023strip}. The motivation behind the proposed Frequency Domain Scale Mixer (FDSM) is to bridge the spatial and frequency domains, enabling layered modulation and feature extraction of global-local characteristics while facilitating intra-domain multi-scale information interaction and fusion. The structure of FDSM is illustrated in Fig. \ref{overview} \textcolor{red}{(e)}. Specifically, in the spatial-scale mixed domain, we first use PWConv to increase the feature dimensions, which are then divided into three groups to extract multi-scale local mixed features. The mathematical formulation is as follows: 
\vspace{-2mm}
\begin{equation}
    X_{Z}=PW(\vartheta (f^{z\times z}(Chunk(PW(X))))),Z\in[ 3,5,7],
    \vspace{-2mm}
\end{equation}
\begin{equation}
    X_{S}=[X_{3},X_{5},X_{7}].
    \vspace{-2mm}
\end{equation}
\begin{table*}
\begin{center}
\caption{Comparison of quantitative results on five datasets. Bold and underlined indicate the best and second-best results.}
\label{table:deraining}
\vspace{-2mm}
\setlength{\tabcolsep}{8pt}
\scalebox{0.8}{
\begin{tabular}{l c c c c c c c c c c c || c c}
\toprule[0.15em]
  & & \multicolumn{2}{c}{\textbf{Test100}~\cite{zhang2019image}}&\multicolumn{2}{c}{\textbf{Rain100H}~\cite{yang2017deep}}&\multicolumn{2}{c}{\textbf{Rain100L}~\cite{yang2017deep}}&\multicolumn{2}{c}{\textbf{Test2800}~\cite{fu2017removing}}&\multicolumn{2}{c||}{\textbf{Test1200}~\cite{zhang2018density}}&\multicolumn{2}{c}{\textbf{Average}}\\
 \textbf{Method} & \textbf{Year} &
 PSNR~$\textcolor{black}{\uparrow}$ & SSIM~$\textcolor{black}{\uparrow}$ & PSNR~$\textcolor{black}{\uparrow}$ & SSIM~$\textcolor{black}{\uparrow}$ & PSNR~$\textcolor{black}{\uparrow}$ & SSIM~$\textcolor{black}{\uparrow}$ & PSNR~$\textcolor{black}{\uparrow}$ & SSIM~$\textcolor{black}{\uparrow}$ & PSNR~$\textcolor{black}{\uparrow}$ & SSIM~$\textcolor{black}{\uparrow}$ & PSNR~$\textcolor{black}{\uparrow}$ & SSIM~$\textcolor{black}{\uparrow}$\\
\midrule[0.15em]
RESCAN~\cite{li2018recurrent} & ECCV2018 & 21.59  & 0.726  & 18.01  & 0.467  & 24.15  & 0.791  & 24.50  & 0.765  & 24.40  & 0.759  & 22.53 & 0.702    \\

PReNet~\cite{ren2019progressive} & CVPR2019 & 23.17  & 0.752  & 17.63  & 0.487  & 27.76  & 0.876  & 27.20  & 0.825  & 26.05  & 0.792  & 24.36 & 0.746    \\

SPDNet~\cite{yi2021structure} & ICCV2021 & 24.25  & 0.848  & 25.87  & 0.809  & 28.63  & 0.880  & 31.05  & 0.904  & 30.42  & 0.893  & 28.04 & 0.867    \\

PCNet~\cite{jiangpcnet} & TIP2021 & 23.29  & 0.762  & 20.83  & 0.563  & 26.64  & 0.817  & 27.10  & 0.818  & 26.53  & 0.791  & 24.88 & 0.750    \\

MPRNet~\cite{Zamir2021MPRNet} & CVPR2021 & 25.66  & 0.859  & 28.23  & 0.850  & 31.94  & 0.930  & 32.14  & 0.925  & 31.32  & 0.901  & 29.86 & 0.893    \\

HINet~\cite{Chen_2021_CVPR} & CVPRW2021 & 23.21  & 0.767  & 20.85  & 0.598  & 27.03  & 0.842  & 28.36  & 0.843  & 27.77  & 0.821  & 25.44 & 0.774    \\

DANet~\cite{jiangDANet2022} & IJCAI2022 & 23.96  & 0.839  & 23.00  & 0.791  & 29.51  & 0.906  & 30.32  & 0.903  & 29.99  & 0.888  & 27.36 & 0.865    \\

Uformer~\cite{Wang_2022_CVPR} & CVPR2022 & 23.87  & 0.815  & 22.43  & 0.700  & 28.39  & 0.883  & 29.71  & 0.886  & 28.65  & 0.856  & 26.61 & 0.828    \\

ALformer~\cite{magic_elf} & ACMMM2022 & 24.41  & 0.844  & 25.10  & 0.807  & 29.39  & 0.903  & 31.36  & 0.916  & 30.40  & 0.897  & 28.13 & 0.874    \\

NAFNet~\cite{chen2022simple} & ECCV2022 & 25.75  & 0.845  & 26.76  & 0.813  & 31.27  & 0.925  & 31.71  & 0.918  & 30.62  & 0.892  & 29.22 & 0.879    \\


MFDNet~\cite{10348527} & TIP2023 & 25.90  & 0.870  & 27.06  & 0.850  & 32.76  & 0.944  & 31.92  & 0.925  & 31.15  & 0.909  & 29.76 & 0.899    \\

HCT-FFN~\cite{chen2023hybrid} & AAAI2023 & 24.86  & 0.847  & 26.70  & 0.819  & 29.94  & 0.906  & 31.46  & 0.915  & 31.23  & 0.901  & 28.84 & 0.878    \\

DRSformer~\cite{Chen_2023_CVPR} & CVPR2023 & 27.86  & \underline{0.885}  & 28.16  & \underline{0.864}  & \underline{34.79}  & \underline{0.954}  & \underline{32.80}  & \textbf{0.931}  & 30.99  & \underline{0.906}  & 30.92 & \underline{0.908}    \\

ChaIR~\cite{cui2023exploring} & KBS2023 & \underline{28.19}  & 0.879  & 28.69  & 0.862  & 34.52  & 0.953  & \textbf{32.85}  & \textbf{0.931}  & 31.30  & 0.903  & \underline{31.11} & 0.906    \\

IRNeXT~\cite{IRNeXT} & ICML2023 & 25.80  & 0.860  & 27.22  & 0.833  & 31.65  & 0.931  & 30.53  & 0.917  & 29.02  & 0.898  & 28.85 & 0.888    \\

OKNet~\cite{cui2024omni} & AAAI2024 & 25.43  & 0.858  & 24.01  & 0.804  & 31.19  & 0.928  & 29.32  & 0.911  & 27.56  & 0.886  & 27.50 & 0.877    \\

AST~\cite{zhou2024AST} & CVPR2024 & 26.07  & 0.859  & 27.40  & 0.833  & 32.03  & 0.932  & 31.65  & 0.921  & 30.69  & 0.897  & 29.57 & 0.889    \\

SFHformer~\cite{SFHformer} & ECCV2024 & 25.67  & 0.856  & 27.25  & 0.832  & 32.97  & 0.944  & 32.27  & 0.925  & \textbf{31.50}  & 0.904  & 29.94 & 0.892    \\

Nerd-rain~\cite{NeRD-Rain} & CVPR2024 & 27.16  & 0.869  & 28.07  & 0.838  & 33.72  & 0.949  & 32.63  & \underline{0.927}  & 30.45  & 0.890  & 30.41 & 0.895    \\


FSNet~\cite{FSNet} & TPAMI2024 & 27.95  & 0.884  & \underline{28.70}  & 0.860  & 34.10  & 0.952  & 32.68  & \textbf{0.931}  & 31.26  & \textbf{0.910}  & 30.94 & \underline{0.908}    \\

\midrule

\textbf{DMSR (Ours)} & -- & \textbf{28.88} & \textbf{0.890} & \textbf{29.41} & \textbf{0.873} & \textbf{35.19} & \textbf{0.957} & 32.50 & \textbf{0.931} & \underline{31.35} & \textbf{0.910} & \textbf{31.47} & \textbf{0.912} \\

\bottomrule[0.15em]
\end{tabular}}

\end{center}
\vspace{-3.0em}
\end{table*}
Here, \(\vartheta(\cdot)\) denotes the GELU activation function. In the frequency domain, the spatial-scale mixed features are first transformed using the fast Fourier transform (FFT) to produce real and imaginary components. The combined frequency features are then passed through a PWConv for modulation, emphasizing and interactively merging beneficial frequency components for restoration. After modulation, the features are passed through the inverse fast Fourier transform (IFFT) to convert the frequency domain features back into the spatial domain. The above process is formally expressed as follows:
\vspace{-2mm}
\begin{equation}
    R,I=FFT(X_{S}),
    \vspace{-2mm}
\end{equation}
\begin{equation}
    \hat{R},\hat{I}=  \phi (BN(PW([\hat{R},\hat{I}  ]))),
    \vspace{-2mm}
\end{equation}
\begin{equation}
    X_{F}=IFFT(\hat{R},\hat{I}  ),
    \vspace{-2mm}
\end{equation}
where \(BN(\cdot)\) represents batch normalization, and \( \phi (\cdot) \) denotes the ReLU activation function. A residual structure is introduced to ensure training stability, resulting in the output of FDSM.

\section{Experiments}

\subsection{Datasets and Metrics}
{\flushleft\textbf{Datasets.}}
We follow the approach of most previous studies to train and validate our model \cite{Zamir2021Restormer,Zamir2021MPRNet,FSNet,Chen_2021_CVPR,magic_elf}. Specifically, we use 13,712 image pairs collected from multiple datasets for training \cite{8099669,7780668,yang2017deep,derain_zhang_2018,zhang2019image}, and validate on five synthetic datasets (Test100 \cite{zhang2019image}, Rain100H \cite{yang2017deep}, Rain100L \cite{yang2017deep}, Test2800 \cite{fu2017removing}, Test1200 \cite{zhang2018density}). Additionally, we use real-world rainy images \cite{wang2019spatial} to further validate the effectiveness of the model in real-world scenarios.
\vspace{-2mm}
{\flushleft\textbf{Evaluation Metrics.}}
Similar to existing computational methods \cite{FSNet,chen2024rethinking}, we use Peak Signal-to-Noise Ratio (PSNR) and Structural Similarity Index (SSIM) to evaluate the de-rain performance. These metrics are calculated based on the Y channel (luminance) in the YCbCr color space.
\vspace{-2mm}
\subsection{Implementation Details}
During training, we use the Adam optimizer with a patch size of \( 64 \times 64 \), a batch size of 12, and a total of 300 epochs. The initial learning rate is set to \( 2 \times 10^{-4} \) and follows a cosine annealing scheduler with linear warm-up for the first 3 epochs. For data augmentation, we follow the same strategies as in previous studies \cite{Zamir2021MPRNet,Kui_2020_CVPR}. The entire framework is implemented in PyTorch and trained on a single NVIDIA GeForce RTX 4090 GPU. During testing, a sliding window slicing method is applied for image cropping \cite{chen2024rethinking}.
\vspace{-2mm}
\subsection{Comparisons with SOTA Methods}
\begin{figure}[t]
	\centering
	\includegraphics[width=0.8\linewidth]{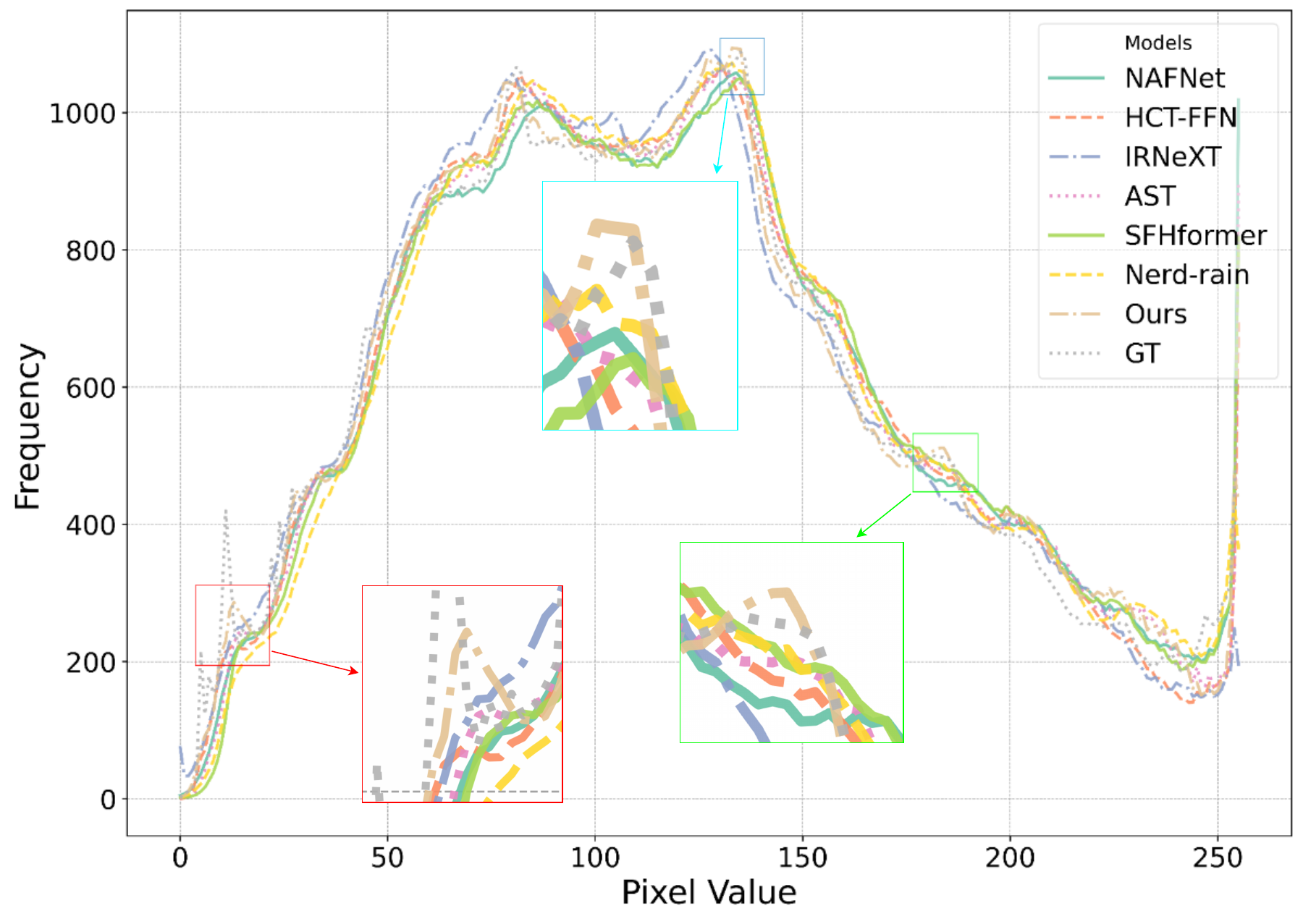} 
	\vspace{-2mm}
	\caption{The average fitting results of the Y channel histogram curve in the YCbCr space on the synthetic dataset.}
	\label{freq}
	\vspace{-3mm}
\end{figure}

\begin{figure*}[!ht]
	\centering
	\includegraphics[width=\linewidth]{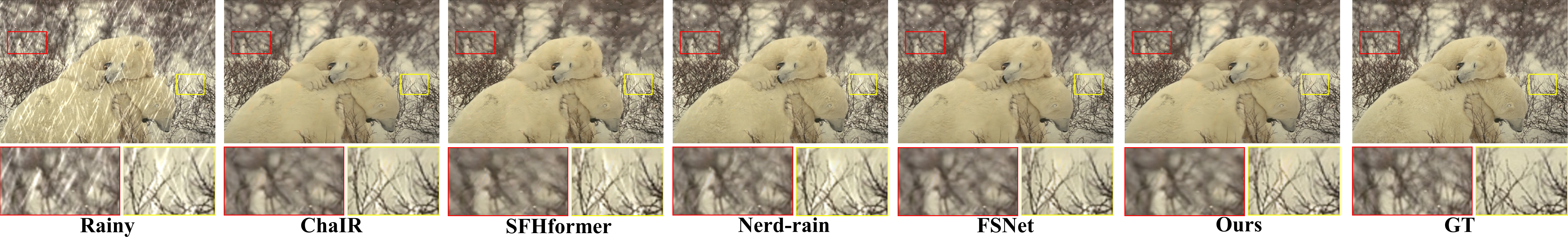} 
	\vspace{-6mm}
	\caption{Visual comparison on the Rain100H dataset \cite{yang2017deep}. Best viewed by zooming in the figures on high-resolution displays.}
	\label{vis1}
	\vspace{-4mm}
\end{figure*}

\begin{figure*}[!ht]
	\centering
	\includegraphics[width=\linewidth]{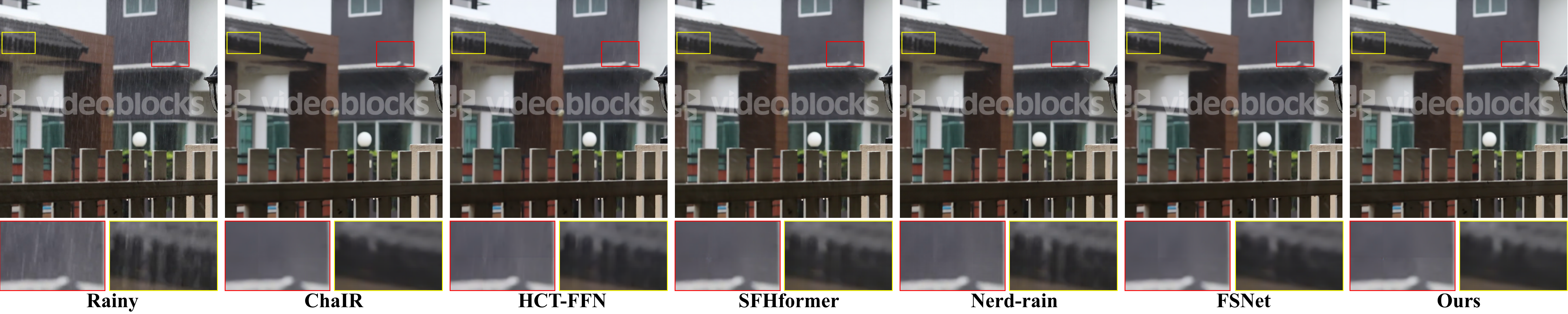} 
	\vspace{-6mm}
	\caption{Visual quality comparison on real-world dataset \cite{wang2019spatial}. Best viewed by zooming in the figures on high-resolution displays.}
	\label{vis2}
	\vspace{-6mm}
\end{figure*}

We compare DMSR with 20 state-of-the-art image deraining methods: RESCAN \cite{li2018recurrent}, PreNet \cite{ren2019progressive}, SPDNet \cite{yi2021structure}, PCNet \cite{jiangpcnet}, MPRNet \cite{Zamir2021MPRNet}, HINet \cite{Chen_2021_CVPR}, DANet \cite{jiangDANet2022}, Uformer \cite{Wang_2022_CVPR}, ALformer \cite{magic_elf}, NAFNet \cite{chen2022simple}, MFDNet \cite{10348527}, HCT-FFN \cite{chen2023hybrid}, DRSformer \cite{Chen_2023_CVPR}, ChaIR \cite{cui2023exploring}, IRNeXT \cite{IRNeXT}, OKNet \cite{cui2024omni}, AST \cite{zhou2024AST}, SFHformer \cite{SFHformer}, Nerd-rain \cite{NeRD-Rain}, and FSNet \cite{FSNet}. To ensure fairness, we retrain these methods from scratch in our environment using their official source codes, without any pretraining or fine-tuning processes.
\vspace{-2mm}
{\flushleft\textbf{Synthetic datasets.}}
Table \ref{table:deraining} presents the comparison results. It is evident that, thanks to the dual-domain multi-scale architecture, DMSR achieves superior performance across five benchmark datasets, notably outperforming the recent state-of-the-art method FSNet \cite{FSNet} by 0.71 dB on Rain100H \cite{yang2017deep}. Additionally, Fig. \ref{vis1} illustrates the visual quality comparison of samples generated by recent methods. Leveraging the advantages of our architecture, DMSR effectively removes rain streaks while preserving reliable background textures, delivering clearer deraining results consistent with the quantitative findings. Furthermore, we present the average fitting results of the Y channel histogram in the YCbCr space on the Rain100H dataset\cite{yang2017deep}, showing that our DMSR deraining results closely match the GT image distribution (see Fig. \ref{freq}).
\vspace{-2mm}
{\flushleft\textbf{Real-world datasets.}}
To validate the performance of DMSR in real-world scenarios, we conducted comparisons on real datasets \cite{wang2019spatial}. As shown in Fig. \ref{vis2}, our DMSR outperforms other sota methods in both rain removal and detail restoration, further demonstrating the advantages of our coarse-to-fine multi-scale paradigm across inter-scale and intra-scale dimensions, as well as the dual-domain high-level feature representation.
\vspace{-2mm}
{\flushleft\textbf{Perceptual quality assessment.}}
To evaluate the perceptual quality of DMSR, we followed the method in \cite{zhou2024AST,Chen_2023_CVPR} and randomly selected 20 rain images from a real-world internet dataset for assessment \cite{wang2019spatial}. As shown in Table \ref{table:niqe}, indicate that DMSR achieves the lowest NIQE, meaning it produces images with better perceptual quality after rain removal.

\vspace{-2mm}
{\flushleft\textbf{More benchmark datasets.}}
The \textcolor{purple}{supplements} present additional experimental results on other benchmark datasets.
\vspace{-2mm}

\begin{table}[t]
\begin{center}
\caption{Results of Perceptual Quality Assessment.}
\label{table:niqe}
\vspace{-2mm}
\setlength{\tabcolsep}{9pt}
\resizebox{1\linewidth}{!}{
\begin{tabular}{cccccccccccccccc}
\toprule[0.8pt]
\multicolumn{2}{l|}{{Methods}} & \multicolumn{2}{c}{Input}   & \multicolumn{2}{c}{DRSformer \cite{Chen_2023_CVPR}} & \multicolumn{2}{c}{SFHformer}\cite{SFHformer} & \multicolumn{2}{c}{Nerd-rain \cite{NeRD-Rain}} &\multicolumn{2}{c}{FSNet \cite{FSNet}}&\multicolumn{2}{|c}{DMSR} \\ 
\midrule[0.8pt]
\multicolumn{2}{l|}{NIQE $\downarrow$} &\multicolumn{2}{c}{5.923}   & \multicolumn{2}{c}{5.814}  & \multicolumn{2}{c}{5.745} & \multicolumn{2}{c}{5.711} & \multicolumn{2}{c}{5.667}& \multicolumn{2}{|c}{\textbf{5.582}}    \\
\bottomrule[0.8pt]
\end{tabular}
}
\end{center}
\vspace{-9mm}
\end{table}

\begin{figure}[!ht]
	\centering
	\includegraphics[width=0.9\linewidth]{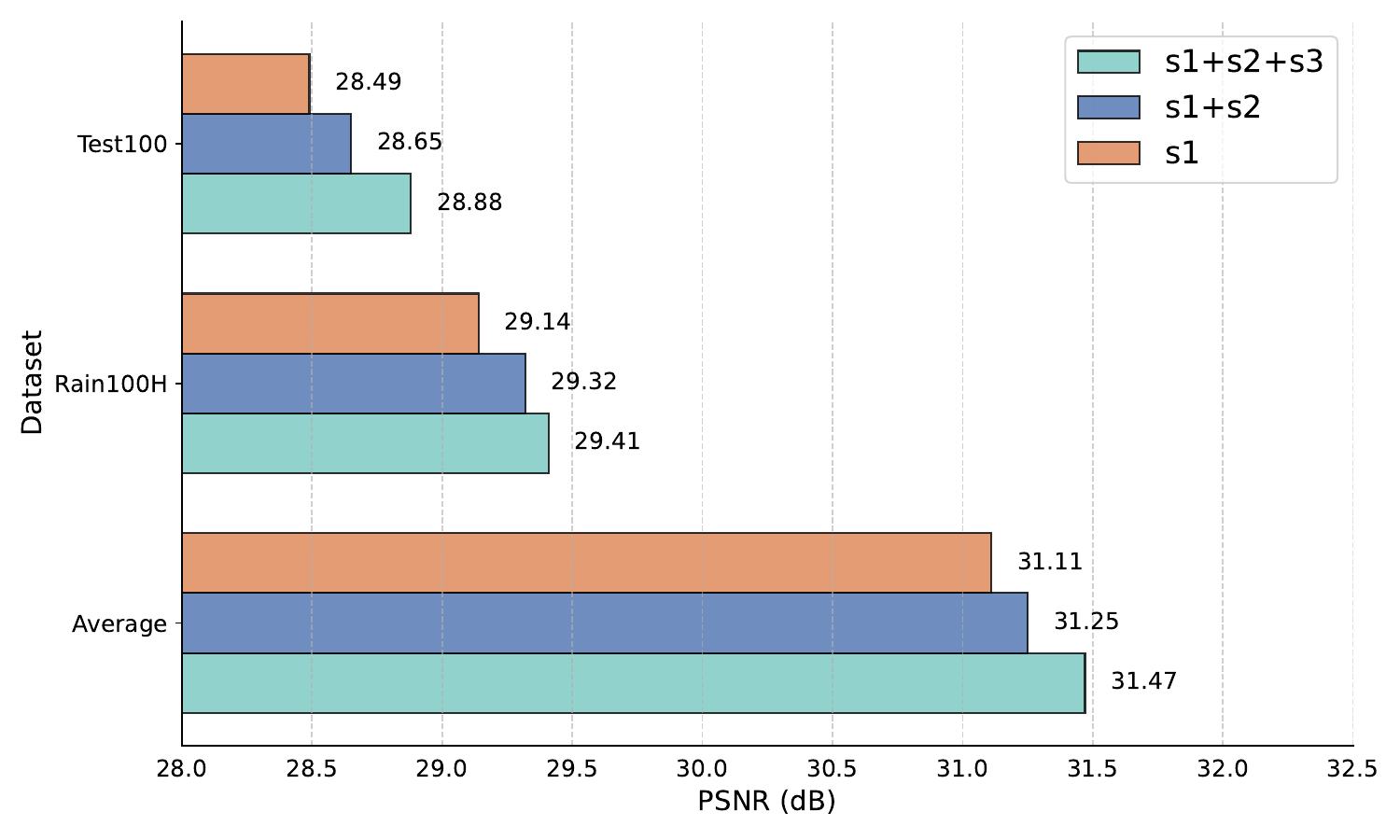} 
	\vspace{-4mm}
	\caption{Ablation analysis of the external multi-scale.}
	\label{abvis1}
	\vspace{-6mm}
\end{figure}
 
\subsection{Ablation Study}
To ensure fairness and consistency, all our ablation experiments were conducted on five synthetic datasets under identical environments and training details. Due to space constraints, we only present results on two datasets and the average results across all five datasets. 

\vspace{-2mm}
{\flushleft\textbf{Effectiveness of external multi-scale.}} 
 To explore the effectiveness of external multi-scale image representations, we compared models with different scales, as shown in Fig. \ref{abvis1} Compared to single-scale rain removal, richer multi-scale representations provide significant gains to the baseline model. The joint effect of coarse and fine scales often leads to higher-quality results.
\vspace{-2mm}
{\flushleft\textbf{Effectiveness of internal multi-scale representation. }}
To further explore the potential of multi-scale architectures, we introduced various forms of internal multi-scale structures in MPSRM and FDSM, creating a synergistic effect with the external multi-scale representations. To validate the effectiveness of internal multi-scale architectures, we progressively removed multi-scale structures within these two components. Specifically, in MPSRM, we denote the branches with \(4 \times \) downsampling and \(2\times\) downsampling as \(scale~branch^{4}\) and \(scale~branch^{2}\), respectively, and remove these branches step by step. For FDSM, we replaced the convolution operations across different scales in the spatial local domain with unified \(3 \times 3\) convolutions or progressively removed these multi-scale convolutions to eliminate its internal multi-scale structure.
\begin{table}[t]
\begin{center}
\caption{Ablation analysis on MPSRM in the intra-scale representation.}
\label{table:ablation2}
\vspace{-2mm}
\setlength{\tabcolsep}{9pt}
\resizebox{1\linewidth}{!}{
\begin{tabular}{l c c  c c c || c c}
\toprule[0.15em]
  & & \multicolumn{2}{c}{\textbf{Test100}}&\multicolumn{2}{c||}{\textbf{Rain100H}}&\multicolumn{2}{c}{\textbf{Average}}\\
 \textcolor{DeepBlack}{\(scale~branch^{4}\)} & \textcolor{DeepBlack}{\(scale~branch^{2}\)} &
 PSNR~$\textcolor{black}{\uparrow}$ & SSIM~$\textcolor{black}{\uparrow}$ & PSNR~$\textcolor{black}{\uparrow}$ & SSIM~$\textcolor{black}{\uparrow}$ & PSNR~$\textcolor{black}{\uparrow}$ & SSIM~$\textcolor{black}{\uparrow}$\\
\midrule[0.1em]
\XSolidBrush & \XSolidBrush  & 28.01  & 0.872  & 29.11  & 0.865  & 30.81 & 0.902    \\

\CheckmarkBold & \XSolidBrush & 28.48  & 0.884  & 29.33  & 0.866  & 31.09 & 0.905    \\

 \XSolidBrush& \CheckmarkBold & 28.53  & 0.887    & 29.36  & 0.868  & 31.14 & 0.907    \\
 \CheckmarkBold &  \CheckmarkBold & 28.88  & 0.890  & 29.41  & 0.873  & 31.47 & 0.912    \\

\bottomrule[0.15em]
\end{tabular}}
\end{center}
\vspace{-6mm}
\end{table}
\begin{table}[th]
\begin{center}
\caption{Ablation analysis on FDSM in the intra-scale representation. The numbers in the table represent the convolution operations with corresponding kernel sizes.}
\label{table:ablation3}
\vspace{-2mm}
\setlength{\tabcolsep}{9pt} 
\resizebox{1\linewidth}{!}{
\begin{tabular}{l c c c c || c c}
\toprule[0.15em]
  & \multicolumn{2}{c}{\textbf{Test100}}&\multicolumn{2}{c||}{\textbf{Rain100H}}&\multicolumn{2}{c}{\textbf{Average}}\\
  \textbf{method} &
 PSNR~$\textcolor{black}{\uparrow}$ & SSIM~$\textcolor{black}{\uparrow}$ & PSNR~$\textcolor{black}{\uparrow}$ & SSIM~$\textcolor{black}{\uparrow}$ & PSNR~$\textcolor{black}{\uparrow}$ & SSIM~$\textcolor{black}{\uparrow}$\\
\midrule[0.1em]
 3  & 28.42  & 0.882  & 29.21  & 0.869  & 31.14 & 0.906    \\

 3+5 & 28.69  & 0.886  & 29.33  & 0.871  & 31.38 & 0.909    \\

 3+5+7 & 28.88  & 0.890    & 29.41  & 0.873  & 31.47 & 0.912    \\
 3+3+3 & 28.55  & 0.884  & 29.30  & 0.869  & 31.29 & 0.910    \\

\bottomrule[0.15em]
\end{tabular}}
\end{center}
\vspace{-9mm}
\end{table}
The results of these comparisons are shown in Tables \ref{table:ablation2} and \ref{table:ablation3}. Removing the internal multi-scale architecture from MPSRM and FDSM consistently led to performance degradation in rain removal. These findings underscore the importance of the dual multi-scale architecture, both external and internal, for achieving high-quality image deraining.
\vspace{-2mm}

{\flushleft\textbf{Ablation Study on MPSRM.}}
We conducted a more detailed study of the individual elements within MPSRM. Specifically, we performed ablation experiments on the SPGA, the skip connections within SPGA, and the \(3 \times 3\) convolution at the end of the module. The results of these experiments are shown in Table \ref{table:ablation4}. When all these critical elements are present, the deraining performance is optimal. This demonstrates the effectiveness of our carefully designed micro-level components.

\vspace{-2mm}
{\flushleft\textbf{Ablation Study on FDSM.}}
We further analyzed the effectiveness of the FDSM in Table \ref{table:ablation5}. The first row demonstrates the significance of spatial local domain operations, highlighting the necessity of local feature processing before extracting global frequency domain features. 
The remaining rows emphasize the importance of mixing global features in the frequency domain. The absence of Fourier transform operations significantly weakens FDSM’s ability to model global dependencies and facilitate frequency domain interactions, leading to a noticeable decline in deraining performance. These results underline the critical role of both local spatial operations and global frequency domain interactions in achieving high-quality deraining.
\vspace{-2mm}

\begin{table}[t]
\begin{center}
\caption{Ablation analysis of the proposed MPSRM.}
\label{table:ablation4}
\vspace{-2mm}
\setlength{\tabcolsep}{9pt}
\resizebox{1\linewidth}{!}{
\begin{tabular}{l c c c  c c c || c c}
\toprule[0.15em]
  & & &\multicolumn{2}{c}{\textbf{Test100}}&\multicolumn{2}{c||}{\textbf{Rain100H}}&\multicolumn{2}{c}{\textbf{Average}}\\
 \textbf{SPGA} & \textbf{Skip connection} & \textbf{3x3 Conv} &
 PSNR~$\textcolor{black}{\uparrow}$ & SSIM~$\textcolor{black}{\uparrow}$ & PSNR~$\textcolor{black}{\uparrow}$ & SSIM~$\textcolor{black}{\uparrow}$ & PSNR~$\textcolor{black}{\uparrow}$ & SSIM~$\textcolor{black}{\uparrow}$\\
\midrule[0.1em]
\XSolidBrush & \XSolidBrush  & \XSolidBrush  & 28.34  & 0.880  & 28.79  & 0.862 & 30.82  &  0.907 \\

\CheckmarkBold & \XSolidBrush & \XSolidBrush  & 28.71  & 0.886  & 29.28  & 0.868 & 31.28  & 0.909 \\

 \CheckmarkBold & \CheckmarkBold &\XSolidBrush & 28.82  & 0.888    & 29.37  & 0.871  & 31.39 & 0.910    \\
 \CheckmarkBold &  \CheckmarkBold & \CheckmarkBold &28.88  & 0.890  & 29.41  & 0.873  & 31.47 & 0.912    \\

\bottomrule[0.15em]
\end{tabular}}
\end{center}
\vspace{-5mm}
\end{table}

\begin{table}[t]
\begin{center}
\caption{Ablation analysis of the proposed FDSM.}
\label{table:ablation5}
\vspace{-2mm}
\setlength{\tabcolsep}{9pt}
\resizebox{1\linewidth}{!}{
\begin{tabular}{l c c c  c c c || c c}
\toprule[0.15em]
  & & &\multicolumn{2}{c}{\textbf{Test100}}&\multicolumn{2}{c||}{\textbf{Rain100H}}&\multicolumn{2}{c}{\textbf{Average}}\\
 \textbf{Multi-Conv} & \textbf{FFT/IFFT} & \textbf{PWConv} &
 PSNR~$\textcolor{black}{\uparrow}$ & SSIM~$\textcolor{black}{\uparrow}$ & PSNR~$\textcolor{black}{\uparrow}$ & SSIM~$\textcolor{black}{\uparrow}$ & PSNR~$\textcolor{black}{\uparrow}$ & SSIM~$\textcolor{black}{\uparrow}$\\
\midrule[0.1em]
\XSolidBrush & \CheckmarkBold & \CheckmarkBold  & 28.29  & 0.878  & 28.83  & 0.867  & 30.89 & 0.905    \\

\CheckmarkBold & \XSolidBrush & \XSolidBrush & 28.41  & 0.881  & 29.02  & 0.869  & 31.02 & 0.908    \\

 \CheckmarkBold & \CheckmarkBold & \XSolidBrush &28.76  & 0.887    & 29.34  & 0.870  & 31.33 & 0.911    \\
 \CheckmarkBold &  \CheckmarkBold  &  \CheckmarkBold & 28.88  & 0.890  & 29.41  & 0.873  & 31.47 & 0.912    \\

\bottomrule[0.15em]
\end{tabular}}
\end{center}
\vspace{-9mm}
\end{table}

\begin{table}[t]
\begin{center}
\caption{Comparison results of the dehazing experiments. See the \textcolor{purple}{supplements} for more results.}
\label{table:dehazing}
\vspace{-2mm}
\setlength{\tabcolsep}{9pt}
\resizebox{1\linewidth}{!}{
\begin{tabular}{l c c c c c c c }
\toprule[0.15em]
  & & \multicolumn{2}{c}{\textbf{RESIDE-6K} \cite{li2018benchmarking}}&\multicolumn{2}{c}{\textbf{Haze-4K} \cite{liu2021synthetic}}&\multicolumn{2}{c}{\textbf{Average}}\\
 \textbf{Method}  & \textbf{Year}  &
 PSNR~$\textcolor{black}{\uparrow}$ & SSIM~$\textcolor{black}{\uparrow}$ & PSNR~$\textcolor{black}{\uparrow}$ & SSIM~$\textcolor{black}{\uparrow}$ & PSNR~$\textcolor{black}{\uparrow}$ & SSIM~$\textcolor{black}{\uparrow}$ \\
\midrule[0.1em]



Dehazeformer~\cite{song2023vision} & TIP2023 & 26.25  & 0.931  & \underline{27.45}  & \underline{0.946}  & 26.85  &   0.939  \\

DEANet \cite{chen2023dea} & TIP2024 & 26.61 & 0.932  & 26.94 & 0.942 & 26.78 & 0.937 \\
SFHfomrer \cite{SFHformer} &  ECCV2024 & \underline{27.08} & \underline{0.940}  & 26.92 & 0.941 & \underline{27.00} &  \underline{0.941} \\
DMSR (ours)  & -- & \textbf{28.56} & \textbf{0.950}  & \textbf{27.64} & \textbf{0.952} & \textbf{28.10} &   \textbf{0.951} \\
\bottomrule[0.15em]
\end{tabular}}
\end{center}
\vspace{-10mm}
\end{table}

\subsection{Other Applications}
To verify whether DMSR can extend its performance to image dehazing, we conducted comparisons on the RESIDE-6K \cite{li2018benchmarking} and Haze-4K \cite{liu2021synthetic} datasets against other dehazing methods, including DEANet \cite{chen2023dea}, SFHformer \cite{SFHformer} and Dehazeformer \cite{song2023vision}. As shown in Table \ref{table:dehazing}, our DMSR achieves the best performance. Meanwhile, we further explored the impact of DMSR's restoration performance on downstream visual tasks, with detailed results provided in the \textcolor{purple}{supplements}.

\section{Conclusion}
In this paper, we propose a novel Dual-Domain Multi-Scale Representation Network (DMSR) for single image deraining. Our architecture integrates both inter-scale and intra-scale information through a collaborative scale representation. The MPSRM removes rain streaks or droplets at multiple scales using a progressive coarse-to-fine approach, while SPGA expands the receptive field by enabling pixel-level perception of extended regions. Additionally, the FDSM extracts multi-scale spatial features and global frequency domain features, enhancing spectral representation through frequency modulation. Extensive experiments on six datasets show that DMSR achieves state-of-the-art performance.

\bibliographystyle{IEEEbib}
\bibliography{references-main}

\end{document}